**Title:** Clinical characteristics, complications and outcomes of critically ill patients with Dengue in Brazil, 2012-2024: a nationwide, multicentre cohort study

*Igor Tona Peres*[1,2],

*Otavio T Ranzani*[3,4],

*Leonardo S. L. Bastos*[1,2],

*Silvio Hamacher*[1,2],

*Tom Edinburgh*[5],

*Esteban Garcia – Gallo*[5],

*Fernando Augusto Bozza*[6,2,7,8]

**Affiliations:**

1 Department of Industrial Engineering (DEI), Pontifical Catholic University of Rio de Janeiro, Rio de Janeiro, Brazil

2 ISARIC, Oswaldo Cruz Foundation (FIOCRUZ), Rio de Janeiro, Brazil

3 Institut de Recerca Sant Pau (IR SANT PAU), Barcelona, Spain

4 Pulmonary Division, Faculty of Medicine, Heart Institute, Hospital das Clínicas da Faculdade de Medicina da Universidade de São Paulo, São Paulo, Brazil

5 ISARIC, Pandemic Sciences Institute, University of Oxford, UK

6 National Institute of Infectious Disease Evandro Chagas (INI), FIOCRUZ, RJ, Brazil

7 D'Or Institute for Research and Education, Rio de Janeiro, Brazil

8 CHRC, NOVA Medical School, Universidade Nova de Lisboa, Lisboa, Portugal

Corresponding authors: Prof. Igor Tona Peres

**Abstract**

**Background.** Dengue outbreaks are a major public health issue, with Brazil reporting 71% of global cases in 2024.

**Purpose.** This study aims to describe the profile of severe dengue patients admitted to Brazilian Intensive Care units (ICUs) (2012–2024), assess trends over time, describe new onset complications while in ICU and determine the risk factors at admission to develop complications during ICU stay.

**Methods.** We performed a prospective study of dengue patients from 253 ICUs across 56 hospitals. We used descriptive statistics to describe the dengue ICU population, logistic regression to identify risk factors for complications during the ICU stay, and a machine learning framework to predict the risk of evolving to complications. Visualisations were generated using ISARIC VERTEX.

**Results.** Of 11,047 admissions, 1,117 admissions (10.1%) evolved to complications, including non-invasive (437 admissions) and invasive ventilation (166), vasopressor (364), blood transfusion (353) and renal replacement therapy (103). Age ≥80 (OR: 3.10, 95% CI: 2.02-4.92), chronic kidney disease (OR: 2.94, 2.22-3.89), liver cirrhosis (OR: 3.65, 1.82-7.04), low platelets (<50,000 cells/mm³; OR: OR: 2.25, 1.89-2.68), and high leukocytes (>7,000 cells/mm³; OR: 2.47, 2.02-3.03) were significant risk factors for complications. A machine learning tool for predicting complications was proposed, showing accurate discrimination and calibration.

**Conclusion.** We described a large cohort of dengue patients admitted to ICUs and identified key risk factors for severe dengue complications, such as advanced age, presence of comorbidities, higher level of leukocytes and lower level of platelets. The proposed prediction tool can be used for early identification and targeted interventions to improve outcomes in dengue-endemic regions.

**Introduction**

Dengue is a mosquito-borne viral infection caused by the dengue virus (DENV) transmitted primarily by *Aedes aegypti,* having four different serotypes and multiple genotypes [1]. Dengue outbreaks are frequent public health events affecting populations and health systems from numerous countries, particularly in tropical and subtropical regions[2]. Currently, the disease is endemic in more than 100 countries. The World Health Organization (WHO) reported over 14.3 million cases and 10,839 deaths in 2024, a sharp increase compared to the 4.6 million cases and 2,400 deaths recorded in 2023[3]. Brazil appears as the most affected country, accounting for 71% of global reported cases and 57% of deaths, with 10.2 million cases and 6,239 deaths in 2024, marking the highest dengue outbreak globally[3].

Regarding severe dengue cases, the WHO reported 52,311 severe cases worldwide in 2024, with Brazil accounting for 8,268 of these cases[3]. Despite the expanding outbreak, comprehensive studies on the clinical characteristics of severe dengue patients remain scarce. Existing literature has primarily focused on sociodemographic factors and general symptoms rather than detailed clinical profiles[4]. Few studies analysed clinical risk factors of severe dengue; however, they used small cohort sizes (<200 patients) to develop the analysis[5,6]. To address this gap, this study aims to analyse the characteristics of severe dengue patients admitted to Intensive Care Units (ICUs) over a 13-year period (2012–2024) within one of Brazil's largest hospital networks. Additionally, it seeks to identify clinical risk factors associated with complications during ICU stays. Severe dengue can lead to critical conditions such as severe plasma leakage, haemorrhage, or organ failure, which may escalate rapidly during the critical phase of illness, requiring urgent intervention to prevent mortality[1].

The key research questions addressed in this study are: (i) What are the demographic and clinical profiles of ICU dengue patients; (ii) How has this profile changed from 2012 to 2024; (iii) How many patients develop complications during ICU stays; and (iv) What are the risk factors at ICU admission for developing complications during ICU stay.

## Materials and Methods

*Study Design*

This is a prospectively study considering all ICU admissions from January 2012 to June 2024. The study was conducted across 253 general medical-surgical ICUs within 56 hospitals from 14 Brazilian states, all part of an integrated hospital network (Rede D'Or-Sao Luiz), and according to the STROBE statement[7]. The study received approval from both the local Ethics Committee and the Brazilian National Ethics Committee (CAAE: 17079119.7.0000.5249) and the need for informed consent was waived due to its retrospective observational design.

*Participants and data collection*

We extracted all anonymized data from an electronic system designed for ICU quality improvement and benchmarking (Epimed Monitor®, Rio de Janeiro, Brazil)[8]. This database includes prospectively collected structured data on all adult ICU admissions (≥16 years old), providing comprehensive clinical information. It captures features collected at admission (first 24h), such as patient demographics, ICU admission diagnoses, comorbidities, illness severity scores, and use of organ support; and the highest or lowest haematological values collected at first 1h of ICU admission. We also have complications and outcomes collected during the complete ICU stay, such as use of devices, ICU and hospital mortality, as well as length of stay (LOS). Table S1 shows the data dictionary including a complete description of each variable comprised in the study. The inclusion criteria were patients with confirmed dengue infection at ICU admission (primary or secondary diagnosis based on laboratory confirmation). In this prospectively collected database, patients are prospectively categorized by a standardized form in "main reason of ICU admission", which reflects the primary admission diagnosis, and for other secondary reasons of ICU admission. We excluded admissions with undefined outcome (42 admissions), readmissions within 30 days (32 admissions), and patients with undefined sex (three admissions) (Figure S1). Since patient sex will be one of the model's confounders, the exclusion of undefined sex is related to the few cases in this category.

*Outcomes*

The primary outcome of interest was the presence of any complication during the ICU stay, defined as a clinically significant event requiring ICU-level intervention and prompting a clear change in management or resource utilization, such as: 1) the need for organ support (e.g., invasive mechanical ventilation, non-invasive ventilation vasopressor drugs, or renal replacement therapy); 2) any blood product transfusion (e.g., red blood cells, platelets and fresh frozen plasma); or (3) death. The secondary outcomes were 60-day ICU and in-hospital mortality, and ICU and in-hospital LOS.

*Statistical analysis*

We used descriptive statistics to describe the ICU population, including demographics, illness severity, comorbidities, complications, laboratory tests, and outcomes. Categorical variables were expressed as frequencies and proportions, while continuous variables were summarized as medians and interquartile ranges (IQR). We included p-values to assess differences between groups (complicated vs not complicated). The Wilcoxon rank sum test was used for continuous variables and the Pearson's Chi-squared test was employed for categorical ones. To illustrate trends over time, we included bar and line plots showing the time series of ICU admissions and outcomes across the years.

To identify risk factors for complications during the ICU stay, we evaluated the association between patient clinical characteristics and the primary outcome. The set of features included in the analysis was selected by clinical relevance. Collinearity among variables was pre-assessed using Pearson's correlation coefficient. Then, we used a univariable logistic regression model to estimate the individual association between each variable and the outcome. We selected the features to the multivariable logistic regression based on clinical judgment and considering the p-value of the univariable regression. Association was calculated by the estimated odds ratios (ORs) and their corresponding 95% confidence intervals (CIs).

We also developed a model to predict the risk of evolving dengue ICU complications following a machine learning framework[9,10]. We used stratified random sampling [11] to split 80% of the dataset for training and 20% for testing. Stratified random sampling selects a sample from a population by dividing the population into subgroups (strata) based on shared characteristics and then randomly selecting samples from each stratum. This ensures that each subgroup is adequately represented in the training and testing sample [12]. We trained and compared the Logistic Regression with two machine learning models (Gradient Boosting Machine and Randon Forests) using the set of risk factors obtained from the previous analysis. We used 10-fold cross–validation to tune the models and evaluated them based on discrimination and calibration measures. Discrimination was measured by the Area Under the Curve (AUC) and calibration was addressed by the calibration belts[13]. Variables with more than 30% missing data were excluded from the multivariable analysis, while the remaining variables were imputed using the Multiple Imputation by Chained Equations (MICE) algorithm[14].

We performed two experiments in terms of prediction modelling: (i) the first one used the same set of features from the previous multivariable analysis; (ii) the second one grouped all comorbidities into one unique feature, called "number of comorbidities", which has three categories (0; 1-2; 3+). We developed an open-source website including the risk calculator that could be used to generate the predictions of complications for ICU dengue patients: https://l4x2nl-igor-peres.shinyapps.io/Dengue_Risk_Calculator/. We also included the model formulation in the results in case other researchers want to apply our model in different contexts.

Plots and figures were generated using ISARIC VERTEX[15], a web-based application that produces tables and visualizations to support key research questions. VERTEX is part of the ISARIC Clinical Epidemiology Platform, developed to standardize and streamline data collection and analysis in outbreak-related clinical research. A public implementation of VERTEX for this study is available at: https://projects.vertex.isaric.org/?param=Dengue_ICU. The developed codes for VERTEX are publicly available at GitHub (https://github.com/ISARICResearch/VERTEX).

**Results**

A total of 11,047 adult patients were included in the analysis. The median age was 45 years (IQR: 32-63) and 91% were from the emergency department (Table 1). The median SAPS-3 was 40 (36-46) and with a median SOFA of 2 (0-3). The 60-day in-hospital mortality was 1.4% (n=157) and the median hospital LOS was 5 (3-7). A total of 1,117 admissions (10.1%) developed complications during ICU stay, such as non-invasive ventilation (n=437; 4.0%), use of vasopressor (n=364; 3.3%), blood transfusion (n=353; 3.2%), invasive mechanical ventilation (n=166; 1.5%), renal replacement therapy (n=103; 0.9%) and high-flow nasal cannula (n=23; 0.2%). Hypertension (n=3,291; 30%), diabetes (n=1,644; 15%) and cardiovascular disease (n=913; 8.3%) were the most prevalent comorbidities. Among the 1,117 patients who developed complications during ICU stay, 364 (33%) required vasopressors or had shock at admission (first 24h), while 311 (28%) required non-invasive ventilation.

Mechanical ventilation and renal replacement therapy were less common at admission, observed in 86 (7.7%) and 60 (5.4%) patients, respectively.

**Table 1.** Profile of ICU dengue admissions

| Feature | Overall N = 11,047 | Any complication during ICU stay | | p-value |
|---|---|---|---|---|
| | | No N = 9,930 (89.9%) | Yes N = 1,117 (10.1%) | |
| **Demographics** | | | | |
| Age, years | 45 (32, 63) | 44 (32, 61) | 58 (38, 74) | <0.001 |
| Sex (Female) | 5,743 (52%) | 5,173 (52%) | 570 (51%) | 0.5 |
| **Comorbidities** | | | | |
| Hypertension | 3,291 (30%) | 2,805 (28%) | 486 (44%) | <0.001 |
| Diabetes | 1,644 (15%) | 1,375 (14%) | 269 (24%) | <0.001 |
| Cardiovascular disease | 913 (8.3%) | 740 (7.5%) | 173 (15%) | <0.001 |
| Malignancy | 458 (4.1%) | 357 (3.6%) | 101 (9.0%) | <0.001 |
| Cerebrovascular disease | 375 (3.4%) | 280 (2.8%) | 95 (8.5%) | <0.001 |
| Asthma | 306 (2.8%) | 271 (2.7%) | 35 (3.1%) | 0.4 |
| Chronic kidney disease | 281 (2.5%) | 183 (1.8%) | 98 (8.8%) | <0.001 |
| Congestive Heart Failure | 198 (1.8%) | 158 (1.6%) | 40 (3.6%) | <0.001 |
| Immunosuppression | 256 (2.3%) | 197 (2.0%) | 59 (5.3%) | <0.001 |
| Obesity | 247 (2.2%) | 197 (2.0%) | 50 (4.5%) | <0.001 |
| Active smoker | 227 (2.1%) | 186 (1.9%) | 41 (3.7%) | <0.001 |
| Liver cirrhosis | 45 (0.4%) | 30 (0.3%) | 15 (1.3%) | <0.001 |
| **Admission** | | | | |
| Admission Source Name | | | | <0.001 |
|   Emergency room | 10,014 (91%) | 9,108 (92%) | 906 (81%) | |
|   Ward/Floor/Step down Unit | 674 (6.1%) | 534 (5.4%) | 140 (13%) | |
|   Other | 355 (3.2%) | 285 (2.9%) | 70 (6.3%) | |
| Modified Frailty Index (MFI) | | | | <0.001 |
|   Non-frail (MFI=0) | 6,274 (57%) | 5,818 (59%) | 456 (41%) | |
|   Pre-frail (MFI=1-2) | 3,888 (35%) | 3,444 (35%) | 444 (40%) | |
|   Frail (MFI>=3) | 885 (8.0%) | 668 (6.7%) | 217 (19%) | |
| Saps-3, points | 40 (36, 46) | 40 (35, 45) | 47 (41, 56) | <0.001 |
| Sofa, points | 2 (0, 3) | 2 (0, 3) | 3 (1, 5) | <0.001 |
| Period | | | | <0.001 |
|   2012-2021 | 1,682 (15%) | 1,418 (14%) | 264 (24%) | |
|   2022-2023 | 2,372 (21%) | 2,124 (21%) | 248 (22%) | |
|   2024 | 6,993 (63%) | 6,388 (64%) | 605 (54%) | |
| **Laboratory evaluation at first admission hour** | | | | |
| Lowest Platelets Count (cells/mm$^3$) | 88,000 (51,000, 140,000) | 88,000 (52,000, 139,000) | 89,000 (37,000, 151,000) | 0.2 |
| Highest Leukocyte Count (cells/mm$^3$) | 3,800 (2,600, 6,100) | 3,700 (2,600, 5,800) | 5,100 (3,200, 9,000) | <0.001 |
| Highest Arterial Lactate (mmol/L) | 1.40 (1.10, 2.00) | 1.40 (1.10, 2.00) | 1.64 (1.20, 2.60) | <0.001 |
| Bun | 11 (8, 15) | 11 (7, 15) | 15 (10, 24) | <0.001 |
| Highest Creatinine | 0.80 (0.64, 1.01) | 0.80 (0.64, 1.00) | 0.90 (0.70, 1.25) | <0.001 |
| Lowest Systolic Blood Pressure | 120 (108, 132) | 120 (108, 131) | 120 (107, 135) | 0.07 |
| Lowest Diastolic Blood Pressure | 71 (63, 80) | 71 (63, 80) | 71 (61, 80) | 0.12 |
| Lowest Mean Arterial Pressure | 88 (79, 96) | 87 (79, 96) | 88 (77, 98) | 0.7 |
| Lowest PaO2FiO2 | 143 (67, 310) | 138 (62, 310) | 157 (100, 292) | 0.003 |
| Lowest Glasgow Coma Scale | 15 (15, 15) | 15 (15, 15) | 15 (15, 15) | <0.001 |
| Highest Heart Rate | 75 (66, 85) | 75 (66, 85) | 78 (68, 89) | <0.001 |
| Highest Respiratory Rate | 18 (16, 20) | 18 (16, 20) | 18 (16, 20) | <0.001 |
| Highest Bilirubin | 0.50 (0.31, 0.70) | 0.50 (0.31, 0.70) | 0.52 (0.34, 0.80) | <0.001 |
| Highest Temperature | 36.2 (36.0, 36.6) | 36.2 (36.0, 36.6) | 36.2 (36.0, 36.7) | 0.7 |
| **Complications at admission** | | | | |
| Vasopressors | 364 (3.3%) | 0 (0%) | 364 (33%) | |
| Noninvasive ventilation | 311 (2.8%) | 0 (0%) | 311 (28%) | |
| Mechanical ventilation | 86 (0.8%) | 0 (0%) | 86 (7.7%) | |
| Renal replacement therapy | 60 (0.5%) | 0 (0%) | 60 (5.4%) | |
| **Complications during ICU stay** | | | | |
| Noninvasive ventilation | 437 (4.0%) | 0 (0%) | 437 (39%) | |
| Vasopressors | 430 (3.9%) | 0 (0%) | 430 (38%) | |
| Transfusion | 353 (3.2%) | 0 (0%) | 353 (32%) | |
| Mechanical ventilation | 166 (1.5%) | 0 (0%) | 166 (15%) | |
| Renal replacement therapy | 103 (0.9%) | 0 (0%) | 103 (9.2%) | |
| High-flow nasal cannula | 23 (0.2%) | 0 (0%) | 23 (2.1%) | |
| **Outcomes** | | | | |
| ICU LOS, days | 3 (2, 4) | 3 (2, 4) | 4 (3, 7) | <0.001 |
| Hospital LOS, days | 5 (3, 7) | 5 (3, 6) | 7 (5, 11) | <0.001 |
| 60-day ICU Mortality | 109 (1.0%) | 12 (0.1%) | 97 (8.7%) | <0.001 |
| 60-day Hospital Mortality | 157 (1.4%) | 37 (0.4%) | 120 (11%) | <0.001 |

We analysed trends in ICU admissions and mortality over time (Figure 1). The number of ICU admissions with a dengue diagnosis was higher in 2024, corresponding to almost half of the cohort size, characterizing a large dengue outbreak. In contrast, we noted a declining trend over the years when analysing the proportion of admissions with any complication and the in-hospital mortality rate. The in-hospital mortality rate was significantly higher ($p < 0.001$) during 2012–2021 (3%) compared to 2022–2023 (0.9%) and 2024 (1.2%). Similarly, the proportion of patients with complications decreased, dropping from 16% in 2012–2021 to 10% in 2022–2023 and further to 8.7% in 2024 (p < 0.001). The complete profile of ICU patients over the years can be assessed in Table S2.

**Figure 1.** a) Admissions and in-hospital mortality rate of ICU dengue patients over time; b) 60-day in-hospital mortality rate; and c) proportion of admission with any complication.

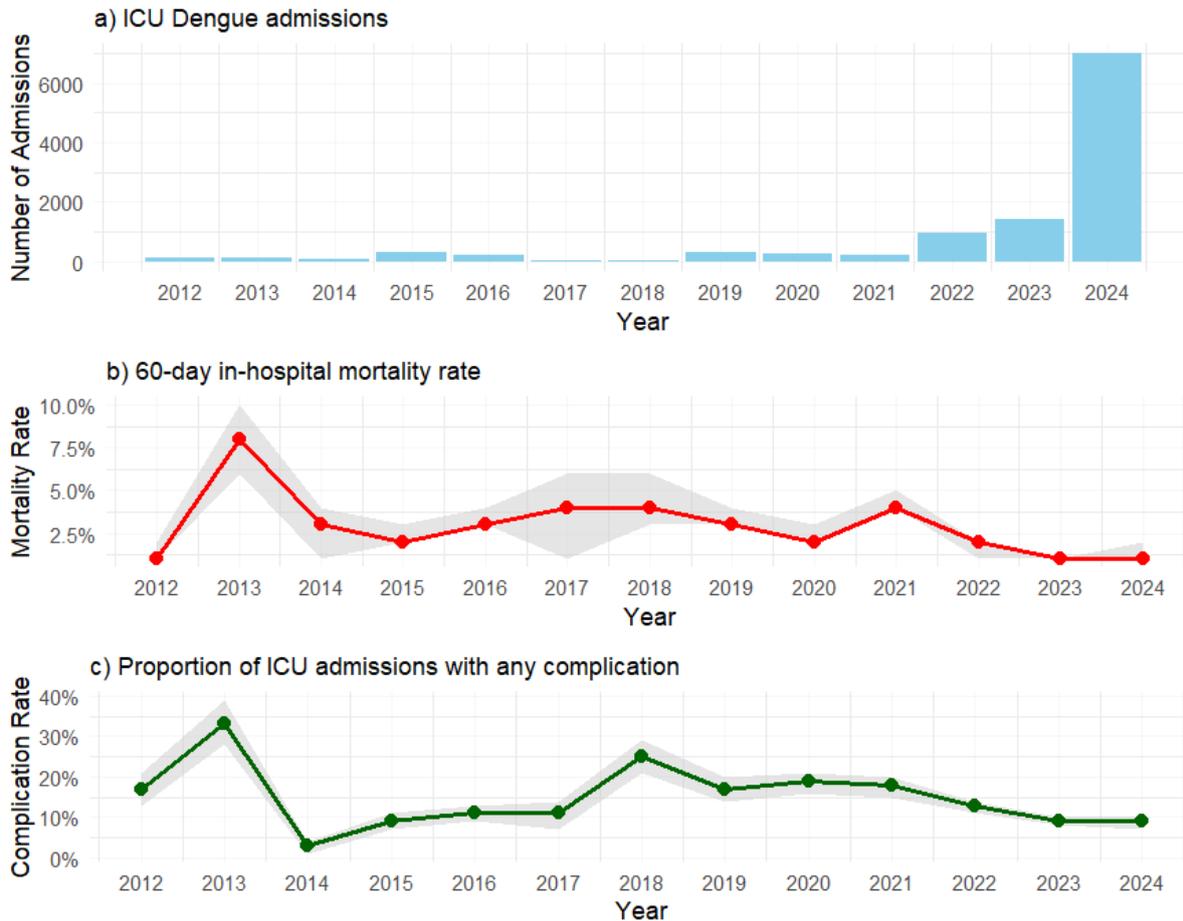

*Complications during ICU Stay*

During the complete ICU stay, vasopressors or shock management were necessary for 430 patients (38%), while non-invasive ventilation was used in 437 patients (39%). Other complications included transfusions (353; 32%), mechanical ventilation (166; 15%), renal replacement therapy (103; 9.2%), and high-flow nasal cannula (23; 2.1%). Patients with complications had longer ICU and hospital LOS (p < 0.001), with a median of 4 (3-7) and 7 (5-11) days, respectively, compared to 3 (2-4) and 5 (3-6) days in patients without complications. Mortality was significantly higher in the former group (p < 0.001), with a 60-day ICU mortality rate of 8.7% and a 60-day hospital mortality rate of 11%, compared to 0.1% and 0.4% in patients without complications. The patients who died were older (75, 59-85), with a higher proportion coming from the ward, floor or step-down unit (16%), presenting a greater number of comorbidities, and had higher severity scores (SAPS-3 = 57, 51-67; SOFA = 4, 2-7; 36% of frail patients). More than half of the patients who required mechanical ventilation (58%) or vasopressor (54%) died, with these two complications being the most impactful on mortality outcomes. The mortality was lower in patients with non-invasive ventilation (27%), transfusion (25%), renal replacement therapy (21%), and high-flow nasal cannula (3.8%).

Patients who developed complications in the ICU showed distinct laboratory values during the first hours of admission compared to those without complications (Table 1). Patients with any complications presented similar median lowest platelet counts at admission compared to those without (median 88,000, IQR: 52,000-139,000; vs. 89,000 IQR: 37,000-151,000 cells/mm³; p = 0.2). However, patients with complications had higher leukocyte count (5,100, 3,200-9,000; vs. 3,700, 2,600-5,800 cells/mm³; $p < 0.001$) and arterial lactate levels (1.64, 1.2-2.6; vs. 1.4, 1.1-2.0 mmol/L; $p < 0.001$) at admission compared to those without complications. Blood pressure and creatinine showed smaller differences but consistently suggested a more severe profile among patients with complications. The lowest systolic, diastolic and mean arterial pressures did not show significant variations. The median age was higher in the complication group [58 (38-74) vs 44 (32-61); $p < 0.001$] and the proportion of female sex was similar [51% vs. 52%; p = 0.5]. All comorbidities were more frequent in the complication group, except asthma (p = 0.4).

In Figure 2, we observe that as age and the number of comorbidities increase, so does the likelihood of developing at least one complication in the ICU (Figure 2b). The proportion of patients developing complications is greater for young adults (16-19 years old) with three or more comorbidities, and we also can note that for all ages the higher the number of comorbidities the greater the proportion of complications (Figure 2b). Patient sex did not appear to be related to the absolute risk of complication (Figure 2a). Additionally, admissions from 2012 to 2023 reported a greater proportion of complications than those from 2024 (Figure 2c). Figures 2d-2f also illustrates the relationship between ICU complications and platelet, leukocyte, and lactate levels (at first 1h of admission), stratified by age groups. Across all ages, lactate levels above 3 mmol/L and leukocyte counts above 7,000 cells/mm³ (leucocytosis) were associated with an increased risk of progression to complications (Figures 2e and 2f). Regarding platelets, although the univariate analysis of Table 1 did not reveal significant effect of platelets count using the continuous value, when we stratify in categories, we can note that levels below 50,000 cells/mm³ (thrombocytopenia) appeared to increase risk across most age groups (Figure 2d). The same effect could be noted when analysing the density distribution of these three laboratory variables (Figure S2). Therefore, we could note that patients with thrombocytopenia, leucocytosis and elevated lactate at baseline (first 1h of ICU admission) were at risk of additional or new complications during ICU stay.

**Figure 2.** Complications by age, sex, number of comorbidities, period of time, platelet levels, leukocyte levels and lactate levels, stratified by age groups.

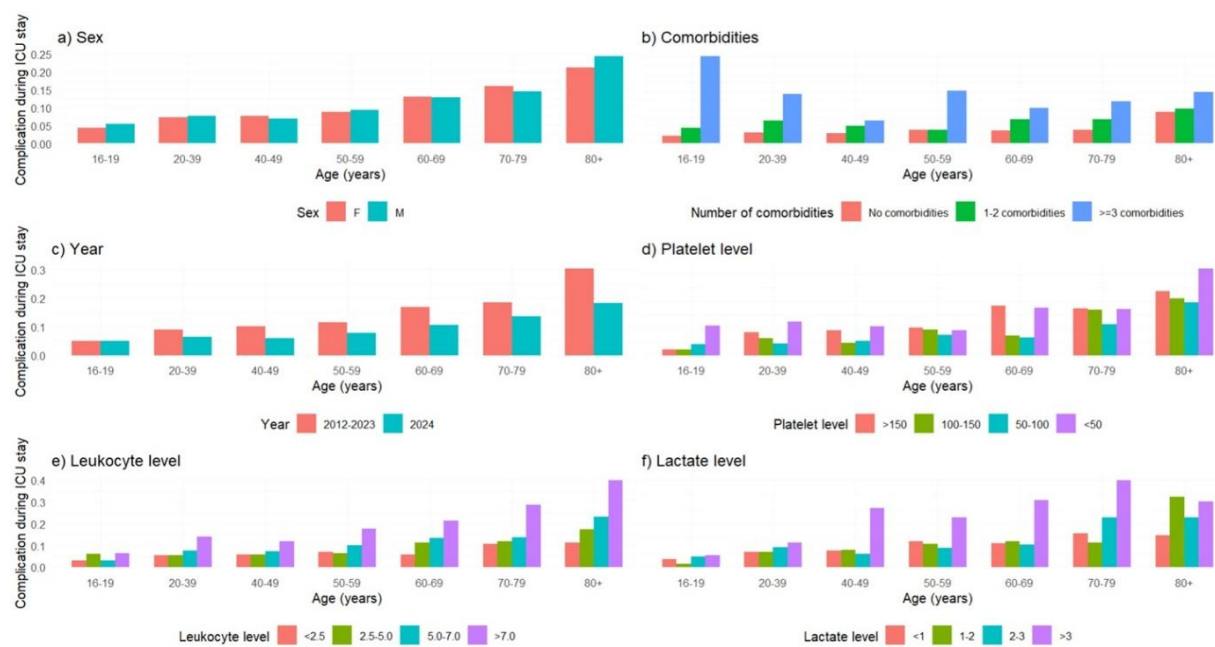

We also analysed the characteristics of these admissions which appears to be in the higher risk group. Compared to the overall population, the patients who presented platelets lower than 50,000 cells/mm³ had greater proportion of new complications (13.6%) and similar age composition (45, 33-61) and slightly higher severity scores (SAPS-3 = 45, 42-50; SOFA = 3, 3-4; 5.6% of frail patients). The most prevalent complication was the blood transfusion (8.8%), followed by vasopressor use or shock (3.2%), non-invasive ventilation (2.6%), mechanical ventilation (1.4%), renal replacement therapy (0.9%), and high-flow nasal cannula (<0.1%). The ICU mortality rate (1.1%) and ICU LOS (3, 2-4) for this subgroup of thrombocytopenia patients were similar to the general population [1.0% and 3(2, 4), respectively].

Regarding the patients who presented leukocyte counts above 7,000 cells/mm³, they presented greater risk of becoming complicated (18.4%) and presented similar age composition (45, 32-65) and severity scores (SAPS-3 = 42, 37-49; SOFA = 2, 0-3; 11% of frail patients). The most prevalent complication was non-invasive ventilation (8.0%) followed by vasopressor use or shock (7.9%), blood transfusion (5.2%), mechanical ventilation (4.6%), renal replacement therapy (2.7%), and high-flow nasal cannula (0.5%). The ICU mortality rate (2.9%) for this subgroup of leucocytosis patients were higher compared to the general population (1.0%); however, the ICU LOS presented similar statistics (3, 2-5).

*Risk factors of ICU complications during ICU Stay*

Table 2 shows the univariable and multivariable logistic regression to investigate the relationship between demographic, clinical, and laboratory variables and the risk of developing complications in the ICU. The features included in the model did not present notable collinearity (Figure S3). We observed in the univariable model that older age [80+, odds ratio (OR): 5.59, 95% CI: 3.77-8.61; p<0.001]; the presence of chronic kidney disease (OR: 5.12,

3.96-6.58; p<0.001), liver cirrhosis (OR: 4.49, 2.35-8.24; p<0.001), cerebrovascular disease (OR: 3.20, 2.51-4.06; p<0.001), obesity (OR: 2.32, 1.67-3.15; p<0.001) and immunosuppression (OR: 2.76, 2.03-3.69; p<0.001) were independently associated with complications. Additionally, low platelet counts (<50,000 cells/mm³) (OR: 2.12, 1.79-2.51; p<0.001), and high leukocyte counts (>7,000 cells/mm³, OR: 3.43, 2.82-4.18; p<0.001) were independently associated with the primary outcome.

**Table 2.** Univariable regression and Multivariable logistic regression to analyse the influence of risk factors of dengue complication

| Feature | Univariable regression OR (95% CI) | p-value | Multivariable Regression OR (95% CI) | p-value |
|---|---|---|---|---|
| **Sex (ref. M)** | | 0.500 | | |
| M | | | | |
| F | 0.96 (0.85, 1.08) | 0.500 | | |
| **Age** | | **<0.001** | | **<0.001** |
| 16-19 (ref.) | | | | |
| 20-39 | 1.55 (1.06, 2.36) | 0.031 | 1.43 (0.97, 2.19) | 0.082 |
| 40-49 | 1.53 (1.03, 2.36) | 0.045 | 1.31 (0.87, 2.03) | 0.200 |
| 50-59 | 1.94 (1.29, 3.00) | 0.002 | 1.56 (1.03, 2.43) | 0.044 |
| 60-69 | 2.85 (1.91, 4.39) | <0.001 | 2.07 (1.36, 3.24) | 0.001 |
| 70-79 | 3.45 (2.32, 5.31) | <0.001 | 2.17 (1.42, 3.43) | <0.001 |
| 80+ | 5.59 (3.77, 8.61) | <0.001 | 3.10 (2.02, 4.92) | <0.001 |
| **Arterial Hypertension** | 1.96 (1.72, 2.22) | **<0.001** | 1.07 (0.91, 1.27) | 0.400 |
| **Diabetes** | 1.97 (1.70, 2.29) | **<0.001** | 1.19 (1.00, 1.42) | **0.047** |
| **Cardiovascular Disease** | 2.28 (1.90, 2.71) | **<0.001** | 1.43 (1.14, 1.78) | **0.002** |
| **Malignancy** | 2.67 (2.11, 3.34) | **<0.001** | 1.66 (1.26, 2.18) | **<0.001** |
| **Cerebrovascular Disease** | 3.20 (2.51, 4.06) | **<0.001** | 1.91 (1.46, 2.49) | **<0.001** |
| **Asthma** | 1.15 (0.79, 1.63) | 0.400 | | |
| **Chronic Kidney Disease** | 5.12 (3.96, 6.58) | **<0.001** | 2.94 (2.22, 3.89) | **<0.001** |
| **Congestive Heart Failure** | 2.30 (1.59, 3.23) | **<0.001** | 0.72 (0.47, 1.10) | 0.140 |
| **Immunosuppression** | 2.76 (2.03, 3.69) | **<0.001** | 1.44 (1.00, 2.05) | **0.043** |
| **Obesity** | 2.32 (1.67, 3.15) | **<0.001** | 2.22 (1.55, 3.08) | **<0.001** |
| **Tobacco Consumption** | 2.00 (1.40, 2.79) | **<0.001** | 1.55 (1.06, 2.21) | **0.020** |
| **Liver Cirrhosis** | 4.49 (2.35, 8.24) | **<0.001** | 3.65 (1.82, 7.04) | **<0.001** |
| **Platelets Class** | | **<0.001** | | **<0.001** |
| 50-100 cells/mm³ (ref.) | | | | |
| >150,000 cells/mm³ | 1.76 (1.47, 2.10) | <0.001 | 1.46 (1.21, 1.76) | <0.001 |
| 100,000-150,000 cells/mm³ | 1.33 (1.10, 1.60) | 0.003 | 1.46 (1.21, 1.76) | <0.001 |
| < 50,000 cells/mm³ | 2.12 (1.79, 2.51) | <0.001 | 2.25 (1.89, 2.68) | <0.001 |
| **Leukocyte Class** | | **<0.001** | | **<0.001** |
| < 2,500 cells/mm³ (ref.) | | | | |
| 2,500-5,000 cells/mm³ | 1.41 (1.16, 1.71) | <0.001 | 1.10 (0.90, 1.35) | 0.400 |
| 5,000-7.000 cells/mm³ | 2.05 (1.63, 2.58) | <0.001 | 1.62 (1.28, 2.05) | <0.001 |
| > 7,000 cells/mm³ | 3.43 (2.82, 4.18) | <0.001 | 2.47 (2.02, 3.03) | <0.001 |

Multivariable logistic regression presents adjusted results by potential confounders. Age remained significantly related to the risk of complications, with older patients (80+) showing a progressively higher odds ratio (OR: 3.10, 2.02-4.92; p<0.001), compared to the reference group (16-19 years). The following comorbidities were significant associated with risk of complications: liver cirrhosis (OR: 3.65, 1.82-7.04; p<0.001), chronic kidney disease (OR: 2.94, 2.22-3.89; p<0.001), obesity (OR: 2.22, 1.55-3.08; p<0.001), cerebrovascular disease (OR: 1.91, 1.46-2.49; p<0.001), malignancy (OR: 1.66, 1.26-2.18; p<0.001), tobacco consumption (OR: 1.55, 1.06-2.21; p=0.02), cardiovascular disease (OR: 1.43, 1.14-1.78; p=0.002), diabetes (OR: 1.19, 1.00-1.42; p=0.047), and immunosuppression (OR: 1.44. 1.00-2.05; p=0.043). Among laboratory variables, lower platelet counts (<50,000 cells/mm³) and elevated leukocyte counts (>7,000 cells/mm³) significantly (p<0.001) increased the odds of complications, with ORs of 2.25 (1.89-2.68) and 2.47 (2.02-3.03), respectively. Figure 3 summarizes the multivariable logistic results. The findings highlight age, specific comorbidities, and laboratory abnormalities as key predictors of dengue complications.

**Figure 3.** Forest plot for the multivariable logistic regression to analyse the influence of risk factors into dengue complication

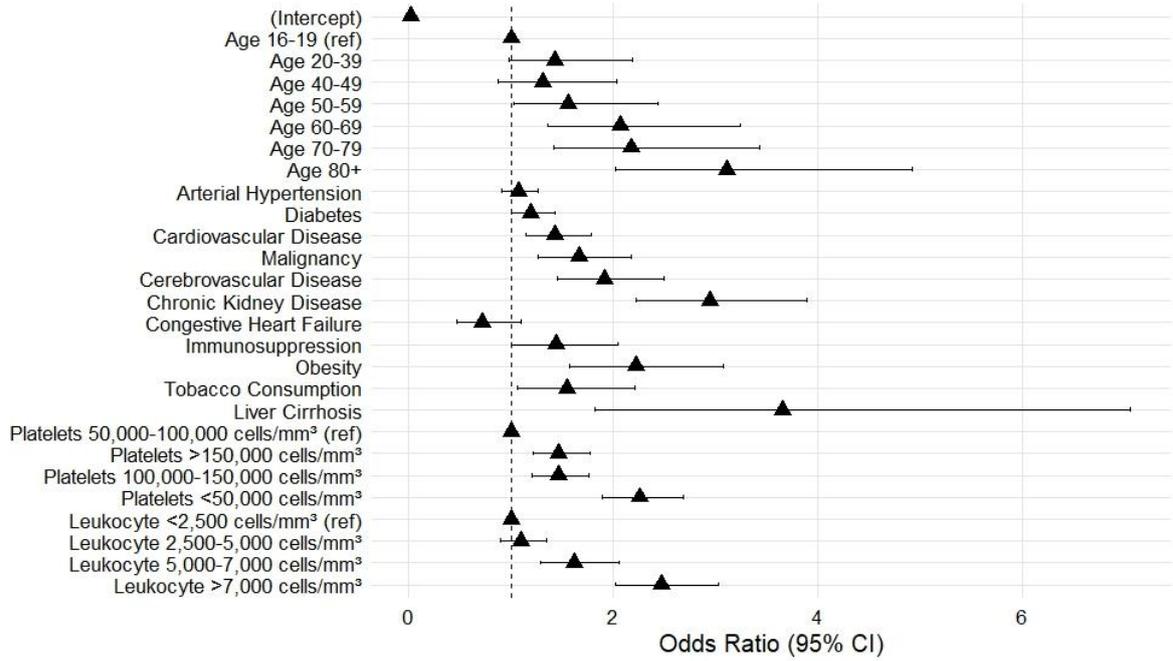

*Prediction model for ICU complications during ICU Stay*

The prediction model using experiment (i) achieved the following results when applied to the testing dataset: Random Forests (AUC = 0.719); GBM (AUC = 0.724); Logistic Regression (AUC = 0.718); and considering experiment (ii): Random Forests (0.711); GBM (0.713); Logistic Regression (0.711). The models presented accurate calibration as can be seen in Figures S4 and S5. To simplify the application process for intensivists, since discrimination and calibration had a small difference between models and between experiments, we propose a simpler model to predict the risk to evolve to dengue complications using the equation (1) of the Logistic Regression model of experiment (ii).

$$\boldsymbol{logit(\hat{p})} = -3.35461 + Age_{class20-39} * 0.31409 + Age_{class40-49} * 0.19208 + Age_{class50-59} * 0.30071 + Age_{class60-69} * 0.51040 + Age_{class70-79} * 0.55154 + Age_{class>80} * 0.97917 + Platelets_{class100-150} * 0.04042 + Platelets_{class50-100} * (-0.29755) + Platelets_{class<50} * 0.44430 + Leukocyte_{class2.5-5} * 0.05018 + Leukocyte_{class5.0-7.0} * 0.52003 + Leukocyte_{class>7.0} * 0.88603 + N_{comorbidities1-2} * 0.53722 + N_{comorbidities3} * 1.23915$$

(1)

**Discussion**

Our study provides a comprehensive analysis of the clinical and demographic profiles of dengue patients admitted to Brazilian ICUs between 2012 and 2024, identifying key risk factors for complications. With a sample size of over 11,000 patients, this represents one of the largest datasets on dengue in critically ill populations.

We identified several significant risk factors for complications in ICU dengue patients, including advanced age, associated comorbidities (chronic kidney disease, liver cirrhosis, obesity, cerebrovascular disease, malignancy, tobacco consumption, cardiovascular disease, diabetes, and immunosuppression) and laboratory markers (low platelet levels and elevated leukocyte levels). These findings are consistent with those of Copaja-Corzo et al.[5] and Carras et al.[6]. Copaja-Corzo et al. [5] identified cardiovascular disease, elevated liver enzymes, and hypofibrinogenemia as key predictors of severe dengue. Comorbidities such as liver cirrhosis are commonly associated with elevated liver enzymes levels. Additionally, comorbidities such as liver cirrhosis, chronic kidney disease and immunosuppression are usually related to hypofibrinogenemia. Carras et al. [6] highlighted the history of cardiovascular disease, the delayed hospital presentation (time from first symptom to hospital consultation over two days) and being of Western European origin as significant predictors for severe dengue.

Our findings on the predictive value of low platelet counts and high leukocyte levels align with those of Trach et al.[16] and Moallemi et al.[17], literature reviews who emphasized the importance of haematological markers, including platelet counts and liver enzymes, in dengue prognosis. However, our study goes further by integrating these biomarkers into a machine learning model to predict complications, achieving an AUC of 0.71. This approach provides a more dynamic and actionable tool for clinicians, particularly in resource-limited settings where early identification of high-risk patients is crucial.

The reviews of Chagas et al.[18] and Díaz-Quijano et al.[19] also identified advanced age, comorbidities (such as diabetes and cardiovascular disease), and organ dysfunction (particularly liver and kidney dysfunction) as significant predictors of mortality and bleeding complications, respectively. These findings are consistent with our results, where low platelet levels and elevated liver enzymes were associated with complications. While Díaz-Quijano et al.[19] specifically focused on bleeding complications, our study aligns with their findings on the importance of platelet counts in predicting severe outcomes. Paraná et al.[4] systematic reviewed and meta-analysed risk factors associated with severe dengue in Latin America and indicated that secondary dengue infection, female sex, white race and specific signs and symptoms were significant predictors of severe dengue outcomes. However, the review identified a lack of studies focusing on clinical characteristics related to severe dengue.

Interestingly, our findings on risk factors for severe dengue complications share similarities with those of Kallas et al.[20] in their study on predictors of mortality in patients with yellow fever. Both studies identified advanced age, liver dysfunction, and renal failure as significant predictors of poor outcomes. However, Kallas et al.[20] also emphasized the role of coagulopathy and elevated bilirubin levels, which were not as prominently featured in our study. This difference may reflect the distinct pathophysiological mechanisms of dengue and yellow fever, with yellow fever being more likely to cause severe hepatic injury and coagulopathy.

Our analysis of trends over time revealed a significant increase in dengue admissions in 2024, coinciding with a global surge in cases reported by the WHO[3]. Despite this increase, we observed a decline in complication and mortality rates, which may reflect improvements in ICU management and early intervention strategies. This finding contrasts with Copaja-Corzo et al.[5], who reported higher mortality rates in their cohort (8.6%), possibly due to differences in healthcare infrastructure, resource availability, and the severity of dengue strains between Brazil and Peru. Other studies have also analysed temporal trends in dengue mortality. Shepard et al.[21] conducted a comprehensive analysis of dengue mortality trends across multiple countries, highlighting significant regional variations in mortality rates. They found that while some regions experienced declining mortality due to improved healthcare access and public health interventions, others saw stable or increasing mortality rates, particularly in areas with limited resources. Similarly, Limkittikul et al.[22] analysed dengue mortality trends in Thailand over a 20-year period, reporting a gradual decline in mortality rates due to better disease management and public health measures. However, they noted periodic spikes in mortality during severe outbreaks.

Our machine learning model, which predicts complications including few features and with an AUC of 0.71, represents a significant advancement in dengue risk stratification. Several other studies have also explored the use of machine learning in dengue prognosis. Most of them analysed the prediction of severe dengue for hospitalised patients or the risk of ICU admission. Torres et al.[23] demonstrated the utility of machine learning in identifying early markers of dengue severity for hospitalised paediatric patients developing a model to predict the risk of ICU admission, while Huang et al.[24] highlighted the importance of combining demographic and laboratory data for accurate risk prediction of severe dengue in hospitalized patients (using WHO 2009 classification). Madewell et al.[25] had the same objective and used clinical and environmental data to predict severe dengue for hospitalised patients, while Sarma et al.[26] explored various machine learning algorithms for dengue prediction for hospitalised patients, showcasing the versatility of these approaches. Our study aligns with the broader trend of using advanced machine learning techniques to improve dengue risk stratification; however, our focus regards to ICU patients. The severe dengue literature lacks in studies analysing critical ill patients. The identification of key risk factors, such as advanced age, chronic comorbidities and abnormal laboratory markers and the use of our proposed model integrated into clinical decision-support systems, which predicts complications with reasonable accuracy, can be used for early identification of high-risk patients in Intensive Care Units. This is particularly relevant in dengue-endemic regions, where timely intervention can significantly reduce mortality and ICU resource utilization.

One of the strengths of our study is the large, nationally representative dataset, which provides granular insights into the critical care context of dengue management. Our machine learning model shows promise, although further validation in external cohorts is needed to ensure its generalizability. We have the following limitations: (i) we analysed a cohort of private ICU patients, which may have different presentation and risk factors than more general populations of dengue patients. However, we showed data from a large network of hospitals with adequate resource availability; and (ii) our definition focuses on objective, clinically significant events—namely, the need for organ support, blood transfusions, and death—which capture complications relevant to ICU-level care. These outcomes partially overlap with the WHO definition of severe dengue[1], which is widely used for triage and early severity stratification in dengue. However, its direct application in the ICU setting is limited, as it was not specifically designed to reflect the clinical course and outcomes of critically ill patients. We believe that clinical interventions (e.g., transfusion) provide a more actionable and meaningful marker of disease severity than laboratory thresholds alone. We acknowledge that this approach may reduce external generalizability. At the same time, we believe it provides a useful framework for characterizing clinical severity in ICU patients and could serve as a foundation for future studies aiming to validate ICU-specific severity definitions in dengue.

Future research should focus on validating our predictive model in diverse settings and exploring the impact of targeted interventions on patient outcomes, such as the ICU datasets of ISARIC hubs in Colombia and Pakistan. Additionally, studies should investigate the role of emerging biomarkers and genetic factors in dengue prognosis, which could further refine risk stratification and treatment strategies. Moreover, future studies should also evaluate the impact of new dengue vaccines approved to be used (e.g., QDenga in Brazil) in the incidence, death and in-hospital mortality.

**Conclusion**

We described a large cohort of dengue patients admitted to ICUs and identified key risk factors for severe dengue complications, such as advanced age, presence of comorbidities, elevated level of leukocytes and low platelets levels. The proposed prediction tool can be used for early identification and targeted interventions to improve outcomes in dengue-endemic regions. By integrating clinical and laboratory data into predictive models, we provide a foundation for more effective and personalized care strategies, which are urgently needed in the face of increasing dengue burden worldwide. As dengue continues to pose a major public health challenge, particularly in endemic regions, the insights from this study can inform both clinical practice and public health strategies to mitigate its impact.

**Funding**

This work was supported by the Brazilian National Council for Scientific and Technological Development (CNPq) [403379/2024-5, 444968/2023-7 and 312654/2023-5 to S.H; 420096/2023-0 to L.B; 309546/2025-7 to I.P]; the Carlos Chagas Filho Foundation for Research Support in Rio de Janeiro State (FAPERJ) [E-26/210.858/2024 and E-26/204.540/2024 to I.P; E-26/204.520/2024 to L.B; E-26/204.187/2024 to S.H]; the Ramón y Cajal program [RYC2023-002923-C to O.R] awarded by the Spanish Ministry of Science; the Innovation and Universities [MICIU/AEI/10.13039/501100011033 to O.R]; the Wellcome Trust [303666/Z/23/Z]; UK International Development [301542-403]; the Gates Foundation [INV-063472]; the European Social Fund Plus (ESF+); the Coordination for the Improvement

**References**

[1] Organization WH. Dengue: Guidelines for Diagnosis, Treatment, Prevention and Control. World Health Organization; 2009.

[2] Cattarino L, Rodriguez-Barraquer I, Imai N, Cummings DAT, Ferguson NM. Mapping global variation in dengue transmission intensity. Sci Transl Med 2020;12:eaax4144. https://doi.org/10.1126/scitranslmed.aax4144.

[3] Global dengue surveillance. World Health Organ 2024. https://worldhealthorg.shinyapps.io/dengue_global/ (accessed August 17, 2024).

[4] Paraná VC, Feitosa CA, da Silva GCS, Gois LL, Santos LA. Risk factors associated with severe dengue in Latin America: A systematic review and meta-analysis. Trop Med Int Health 2024;29:173–91. https://doi.org/10.1111/tmi.13968.

[5] Copaja-Corzo C, Flores-Cohaila J, Tapia-Sequeiros G, Vilchez-Cornejo J, Hueda-Zavaleta M, Vilcarromero S, et al. Risk factors associated with dengue complications and death: A cohort study in Peru. PLOS ONE 2024;19:e0305689. https://doi.org/10.1371/journal.pone.0305689.

[6] Carras M, Maillard O, Cousty J, Gérardin P, Boukerrou M, Raffray L, et al. Associated risk factors of severe dengue in Reunion Island: A prospective cohort study. PLoS Negl Trop Dis 2023;17:e0011260. https://doi.org/10.1371/journal.pntd.0011260.

[7] Elm E von, Altman DG, Egger M, Pocock SJ, Gøtzsche PC, Vandenbroucke JP. The Strengthening the Reporting of Observational Studies in Epidemiology (STROBE) statement: guidelines for reporting observational studies. The Lancet 2007;370:1453–7. https://doi.org/10.1016/S0140-6736(07)61602-X.

[8] Soares M, Borges LP, Bastos L dos SL, Zampieri FG, Miranda GA, Kurtz P, et al. Update on the Epimed Monitor Adult ICU Database: 15 years of its use in national registries, quality improvement initiatives and clinical research. Crit Care Sci 2024;36.

[9] Peres IT, Hamacher S, Cyrino Oliveira FL, Bozza FA, Salluh JIF. Data-driven methodology to predict the ICU length of stay: A multicentre study of 99,492

admissions in 109 Brazilian units. Anaesth Crit Care Pain Med 2022;41:101142. https://doi.org/10.1016/j.accpm.2022.101142.

[10] Peres IT, Hamacher S, Oliveira FLC, Bozza FA, Salluh JIF. Prediction of intensive care units length of stay: A concise review. Rev Bras Ter Intensiva 2021;33:183–7. https://doi.org/10.5935/0103-507X.20210025.

[11] Kim J-H. Estimating classification error rate: Repeated cross-validation, repeated hold-out and bootstrap. Comput Stat Data Anal 2009;53:3735–45. https://doi.org/10.1016/j.csda.2009.04.009.

[12] Kuhn M, Johnson K. Data Pre-processing. In: Kuhn M, Johnson K, editors. Appl. Predict. Model., New York, NY: Springer; 2013, p. 27–59. https://doi.org/10.1007/978-1-4614-6849-3_3.

[13] Walsh CG, Sharman K, Hripcsak G. Beyond discrimination: A comparison of calibration methods and clinical usefulness of predictive models of readmission risk. J Biomed Inform 2017;76:9–18. https://doi.org/10.1016/j.jbi.2017.10.008.

[14] White IR, Royston P, Wood AM. Multiple imputation using chained equations: Issues and guidance for practice. Stat Med 2011;30:377–99. https://doi.org/10.1002/sim.4067.

[15] Garcia-Gallo E, Edinburgh T, Bastos L, Peres I, Imtiaz H, Raffaini LE, et al. ISARIC VERTEX 2024. https://doi.org/10.5281/zenodo.14170825.

[16] Thach TQ, Eisa HG, Hmeda AB, Faraj H, Thuan TM, Abdelrahman MM, et al. Predictive markers for the early prognosis of dengue severity: A systematic review and meta-analysis. PLoS Negl Trop Dis 2021;15:e0009808. https://doi.org/10.1371/journal.pntd.0009808.

[17] Moallemi S, Lloyd AR, Rodrigo C. Early biomarkers for prediction of severe manifestations of dengue fever: a systematic review and a meta-analysis. Sci Rep 2023;13:17485. https://doi.org/10.1038/s41598-023-44559-9.

[18] Chagas GCL, Rangel AR, Noronha LM, Veloso FCS, Kassar SB, Oliveira MJC, et al. Risk factors for mortality in patients with dengue: A systematic review and meta-analysis. Trop Med Int Health 2022;27:656–68. https://doi.org/10.1111/tmi.13797.

[19] Díaz-Quijano FA. Predictors of spontaneous bleeding in dengue patients: a systematic review of the literature. Investig Clínica 2008;49:111–22.

[20] Kallas EG, D'Elia Zanella LGFAB, Moreira CHV, Buccheri R, Diniz GBF, Castiñeiras ACP, et al. Predictors of mortality in patients with yellow fever: an observational cohort study. Lancet Infect Dis 2019;19:750–8. https://doi.org/10.1016/S1473-3099(19)30125-2.

[21] Shepard DS, Undurraga EA, Halasa YA, Stanaway JD. The global economic burden of dengue: a systematic analysis. Lancet Infect Dis 2016;16:935–41. https://doi.org/10.1016/S1473-3099(16)00146-8.

[22] Limkittikul K, Brett J, L'Azou M. Epidemiological Trends of Dengue Disease in Thailand (2000–2011): A Systematic Literature Review. PLoS Negl Trop Dis 2014;8:e3241. https://doi.org/10.1371/journal.pntd.0003241.

[23] Caicedo-Torres W, Paternina Á, Pinzón H. Machine Learning Models for Early Dengue Severity Prediction. In: Montes y Gómez M, Escalante HJ, Segura A, Murillo J de D, editors. Adv. Artif. Intell. - IBERAMIA 2016, Cham: Springer International Publishing; 2016, p. 247–58. https://doi.org/10.1007/978-3-319-47955-2_21.

[24] Huang S-W, Tsai H-P, Hung S-J, Ko W-C, Wang J-R. Assessing the risk of dengue severity using demographic information and laboratory test results with machine learning. PLoS Negl Trop Dis 2020;14:e0008960. https://doi.org/10.1371/journal.pntd.0008960.

[25] Madewell ZJ, Rodriguez DM, Thayer MB, Rivera-Amill V, Paz-Bailey G, Adams LE, et al. Machine learning for predicting severe dengue in Puerto Rico. Infect Dis Poverty 2025;14:5. https://doi.org/10.1186/s40249-025-01273-0.

[26] Sarma D, Hossain S, Mittra T, Bhuiya MdAM, Saha I, Chakma R. Dengue Prediction using Machine Learning Algorithms. 2020 IEEE 8th R10 Humanit. Technol. Conf. R10-HTC, 2020, p. 1–6. https://doi.org/10.1109/R10-HTC49770.2020.9357035.

# Supplementary Material

**Table S1. Data dictionary including a complete description of each variable comprised in the study**

| Field | Name | Description |
|---|---|---|
| Age | Current Age | Current age of the patient in full years |
| Sex | Sex | Patient sex assignment |
| Modified Frailty Index (MFI) | Modified Frailty Index (MFI) (points) | Sum of points automatically calculated from the completion of the variables that score for the Modified Frailty Index - Automatic calculation performed by EPIMED |
| Saps-3, points | SAPS 3 Points | Field automatically calculated by EPIMED through SAPS3 variables |
| Sofa, points | SOFA points | Field automatically calculated by EPIMED through SOFA variables |
| Period | Period | Category related to the year of ICU admission (2012-2021; 2022-2023; 2024) |
| Admission Source Name | Origin of the patient at admission to the unit | Origin of the patient at admission to the unit. Represent the place where the patient came from before admission to the unit |
| Congestive Heart Failure | Heart failure (NYHA - classes II-III-IV) | Previous diagnosis of congestive heart failure according to the NYHA classification and with the need for pharmacological treatment for control; important limitation of physical activity; comfortable at rest, but small physical activities trigger symptoms; inability to perform any physical activity without discomfort; symptoms of heart failure or angina may be present even at rest; any physical activity results in increased discomfort |
| Chronic kidney disease | Chronic kidney disease with or without dialysis | Chronic kidney disease without prior need for any method of renal function replacement; end-stage chronic kidney disease requiring dialysis treatment by hemodialysis or other method of renal function replacement |
| Liver cirrhosis | Cirrhosis (Child A-B-C) | Biopsy-proven cirrhosis and documented portal hypertension; past episodes of gastrointestinal bleeding attributed to portal hypertension; refractory ascites, previous episodes of liver failure, encephalopathy, or coma |
| Malignancy | Hematologic neoplasm; Solid tumor (locoregional or metastatic) | Proven diagnosis of malignant hematological neoplasm (leukemias, lymphomas, myeloproliferative diseases or other hematological disease that require treatment with radiotherapy or chemotherapy for its treatment); diagnosis of solid tumor (proven by histopathological examination) restricted to the organ of origin or with extension to contiguous structures or regional lymph nodes; diagnosis of solid tumor (proven by histopathological examination) with distant metastases, imaging findings (computed tomography, ultrasound, and magnetic resonance imaging) that are highly suggestive of distant metastases in patients with a proven diagnosis of cancer may be considered. |
| **Immunosuppression** | Immunosuppression | Use of immunosuppressive treatments, including corticosteroids in the form of "pulses" or with a dose equal to or greater than 0.3 mg/kg/day of prednisone or equivalent |
| **Hypertension** | Hypertension | Previous diagnosis of systemic arterial hypertension requiring pharmacological treatment |
| **Asthma** | Asthma | Previous diagnosis of bronchial asthma requiring pharmacological treatment for control or prevention |
| **Diabetes** | Diabetes with or without complication | Previous diagnosis of type I or II Diabetes Mellitus requiring pharmacological treatment combined or not with hypoglycemic agents and insulin in a patient with or without clinical or laboratory evidence of disease-related complications (e.g. retinopathy, nephropathy, micro and macrovascular complications) |
| **Cardiovascular disease** | Angina, previous AMI, other cardiac arrhythmias, deep venous thrombosis, chronic atrial fibrillation or peripheral arterial disease | Intermittent claudication, bypass surgery or amputation for gangrene or arterial insufficiency, thoracic/abdominal aortic aneurysm with or without surgery to correct, severe chronic arterial insufficiency; previous diagnosis of acute myocardial infarction occurring at any time prior to surgery; previous diagnosis of bradi or cardiac tachyarrhythmias requiring pharmacological treatment for rhythm maintenance or heart rate control or implantation of a pacemaker or defibrillator; previous diagnosis of pulmonary embolism and/or deep vein thrombosis in any topography in the last six months or need for anticoagulation for prophylaxis of new thromboembolic episodes; cardiac arrhythmia, characterized by a fast, irregular heartbeat; or intermittent claudication, bypass surgery or amputation for gangrene or arterial insufficiency, thoracic/abdominal aortic aneurysm with or without surgery to correct, severe chronic arterial insufficiency. |
| **Cerebrovascular disease** | Stroke with or without sequelae, or dementia | Previous diagnosis of ischemic or hemorrhagic stroke that has or has not resulted in apparent neurological sequelae. The sequelae can be motor, sensory or cognitive; or previous diagnosis of dementia syndrome of any etiology |
| **Active Smoker** | Smoking (within the last 12 months) | Smoking with nicotine dependence in the 12 months prior to current hospital admission |
| **Obesity** | Morbid obesity | Previous diagnosis of morbid obesity defined as follows: 50 kg or more above ideal body weight or having a Body Mass Index (BMI) equal to or greater than 40. This includes people with a history of bariatric surgery |
| **Lowest Systolic Blood Pressure** | Lowest Systolic Blood Pressure | Lowest Systolic Blood Pressure (first 1 hour of admission) |
| **Lowest Diastolic Blood Pressure** | Lowest Diastolic Blood Pressure | Lowest Diastolic Blood Pressure (first 1 hour of admission) |
| **Lowest Mean Arterial Pressure** | Lowest Mean Arterial Pressure | Lowest Mean Arterial Pressure (first 1 hour of admission) |
| **Lowest Glasgow Coma Scale** | Lowest Glasgow Coma Scale | Lowest Glasgow Coma Scale (first 1 hour of admission) |
| **Lowest Platelets Count** | Lowest Platelets Count (cells/mm³) | Lowest Platelets Count (cells/mm³) (first 1 hour of admission) |
| **Lowest PaO2FiO2** | Lowest PaO2FiO2 | Lowest PaO2FiO2 (first 1 hour of admission) |
| **Highest Heart Rate** | Highest Heart Rate | Highest Heart Rate (first 1 hour of admission) |
| **Highest Respiratory Rate** | Highest Respiratory Rate | Highest Respiratory Rate (first 1 hour of admission) |
| **Highest Temperature** | Highest Temperature | Highest Temperature (first 1 hour of admission) |
| **Highest Leukocyte Count** | Highest Leukocyte Count (cells/mm³) | Highest Leukocyte Count (cells/mm³) (first 1 hour of admission) |
| **Highest Creatinine** | Highest Creatinine | Highest Creatinine (first 1 hour of admission) |
| **Highest Bilirubin** | Highest Bilirubin | Highest Bilirubin (first 1 hour of admission) |
| **Highest Arterial Lactate** | Highest Arterial Lactate (mmol/L) | Highest Arterial Lactate (mmol/L) (first 1 hour of admission) |
| **Bun** | Bun | Bun (first 1 hour of admission) |
| **Noninvasive ventilation** | Noninvasive ventilation | Use of noninvasive ventilation during ICU stay (e.g., CPAP - Continuous Positive Airway Pressure, BiPAP - Bilevel Positive Airway Pressure) |
| **Vasopressors** | Vasopressors | Use of vasopressor drugs during ICU stay |
| **Transfusion** | Transfusion | Any blood product transfusion (e.g., red blood cells, platelets and fresh frozen plasma) |
| **Mechanical ventilation** | Mechanical ventilation | Use of mechanical ventilation during ICU stay |
| **Renal replacement therapy** | Renal replacement therapy | Use of renal replacement therapy during ICU stay |
| **High-flow nasal cannula** | High-flow nasal cannula | Use of high-flow nasal cannula during ICU stay |
| **Complication** | Any complication during ICU stay | Use of noninvasive ventilation, vasopressors, transfusion, mechanical ventilation, renal replacement therapy or high-flow nasal cannula during ICU stay |
| **ICU LOS** | ICU LOS, days | Total length of stay in the ICU (days) |
| **Hospital LOS** | Hospital LOS, days | Total length of stay in the hospital (days) |
| **60-day ICU Mortality** | 60-day ICU Mortality | ICU mortality (up to 60 days of ICU admission) |
| **60-day Hospital Mortality** | 60-day Hospital Mortality | Hospital mortality (up to 60 days of hospital admission) |

**Table S2. ICU profile for each period (Overall, 2012-2021, 2022-2023 and 2024)**

| Feature | Overall<br>N = 11,047 | 2012-2021<br>N = 1,682 | 2022-2023<br>N = 2,372 | 2024<br>N = 6,993 | p-value |
|---|---|---|---|---|---|
| **Demographics** | | | | | |
| Age, years | 45 (32, 63) | 46 (32, 63) | 43 (31, 60) | 46 (33, 64) | <0.001 |
| Sex (Female) | 5,743 (52%) | 846 (50%) | 1,180 (50%) | 3,717 (53%) | 0.005 |
| **Comorbidities** | | | | | |
| Hypertension | 3,291 (30%) | 509 (30%) | 646 (27%) | 2,136 (31%) | 0.009 |
| Diabetes | 1,644 (15%) | 225 (13%) | 303 (13%) | 1,116 (16%) | <0.001 |
| Cardiovascular disease | 912 (8.3%) | 111 (6.9%) | 198 (8.3%) | 603 (8.6%) | 0.072 |
| Malignancy | 458 (4.2%) | 53 (3.3%) | 101 (4.3%) | 304 (4.3%) | 0.200 |
| Cerebrovascular disease | 374 (3.4%) | 54 (3.3%) | 60 (2.5%) | 260 (3.7%) | 0.022 |
| Asthma | 306 (2.8%) | 36 (2.1%) | 58 (2.4%) | 212 (3.0%) | 0.075 |
| Chronic kidney disease | 281 (2.6%) | 35 (2.2%) | 72 (3.0%) | 174 (2.5%) | 0.200 |
| Congestive Heart Failure | 198 (1.8%) | 29 (1.7%) | 54 (2.3%) | 115 (1.6%) | 0.130 |
| Immunosuppression | 256 (2.3%) | 44 (2.6%) | 68 (2.9%) | 144 (2.1%) | 0.053 |
| Obesity | 247 (2.2%) | 47 (2.9%) | 60 (2.5%) | 140 (2.0%) | 0.049 |
| Active smoker | 227 (2.1%) | 59 (3.5%) | 39 (1.6%) | 129 (1.8%) | <0.001 |
| Liver cirrhosis | 45 (0.4%) | 16 (1.0%) | 14 (0.6%) | 15 (0.2%) | <0.001 |
| **Admission** | | | | | |
| Admission Source Name | | | | | <0.001 |
|   Emergency room | 10,014 (91%) | 1,465 (87%) | 2,170 (91%) | 6,379 (91%) | |
|   Ward/Floor/Step down Unit | 674 (6.1%) | 161 (9.6%) | 124 (5.2%) | 389 (5.6%) | |
|   Other | 355 (3.2%) | 55 (3.3%) | 78 (3.3%) | 222 (3.2%) | |
| Modified Frailty Index (MFI) | | | | | <0.001 |
|   Non-frail (MFI=0) | 5,761 (55%) | 386 (35%) | 1,412 (60%) | 3,963 (57%) | |
|   Pre-frail (MFI=1-2) | 3,812 (36%) | 564 (52%) | 766 (32%) | 2,482 (35%) | |
|   Frail (MFI>=3) | 884 (8.5%) | 142 (13%) | 194 (8.2%) | 548 (7.8%) | |
| Saps-3, points | 40 (36, 46) | 40 (36, 47) | 40 (35, 45) | 40 (36, 46) | 0.010 |
| Sofa score, points | 2 (0, 3) | 2 (0, 3) | 2 (0, 3) | 2 (0, 3) | <0.001 |
| **Laboratory evaluation at first admission hour** | | | | | |
| Lowest Platelets Count | 87,000 (51,000, 141,000) | 88,000 (46,000, 149,000) | 90,000 (51,000, 157,000) | 86,000 (52,000, 136,000) | <0.001 |
|   Not informed | 871 | 154 | 195 | 522 | |
| Highest Leukocyte Count | 3,800 (2,500, 6,200) | 4,500 (2,800, 7,300) | 4,100 (2,600, 7,000) | 3,600 (2,500, 5,600) | <0.001 |
|   Not informed | 884 | 166 | 199 | 519 | |
| Highest Arterial Lactate | 1.40 (1.10, 2.00) | 1.50 (1.10, 2.30) | 1.43 (1.10, 2.00) | 1.40 (1.10, 2.00) | 0.093 |
|   Not informed | 9.241 | 1,301 | 1,758 | 6.182 | |
| Bun | 11 (8, 15) | 11 (8, 17) | 11 (8, 15) | 11 (7, 15) | <0.001 |
|   Not informed | 1.318 | 362 | 229 | 727 | |
| Creatinine | 0.80 (0.64, 1.01) | 0.88 (0.69, 1.08) | 0.85 (0.70, 1.10) | 0.80 (0.60, 1.00) | <0.001 |
|   Not informed | 1.151 | 238 | 223 | 690 | |
| Lowest Systolic Blood Pressure | 120 (108, 132) | 120 (109, 134) | 119 (107, 131) | 120 (108, 131) | <0.001 |
|   Not informed | 99 | 39 | 34 | 26 | |
| Lowest Diastolic Blood Pressure | 71 (63, 80) | 72 (63, 80) | 71 (62, 80) | 71 (64, 80) | 0.033 |
|   Not informed | 105 | 40 | 34 | 31 | |
| Lowest Mean Arterial Pressure | 88 (79, 96) | 260 (81, 455) | 229 (81, 310) | 88 (79, 96) | 0.002 |
|   Not informed | 105 | 1,438 | 2,097 | 31 | |
| Highest PaO2FiO2 | 143 (67, 310) | 89 (80, 97) | 87 (78, 96) | 104 (61, 262) | <0.001 |
|   Not informed | 9.938 | 40 | 34 | 6.403 | |
| Lowest Glasgow Coma Scale | 15 (15, 15) | 15 (15, 15) | 15 (15, 15) | 15 (15, 15) | 0.200 |
|   Not informed | 294 | 131 | 67 | 96 | |
| Highest Heart Rate | 75 (66, 85) | 78 (67, 89) | 77 (67, 86) | 75 (65, 84) | <0.001 |
|   Not informed | 150 | 41 | 41 | 68 | |
| Highest Respiratory Rate | 18 (16, 20) | 18 (16, 20) | 18 (16, 20) | 18 (16, 20) | <0.001 |
|   Not informed | 113 | 55 | 34 | 24 | |
| Highest Bilirubin | 0.50 (0.31, 0.70) | 0.41 (0.28, 0.64) | 0.47 (0.30, 0.70) | 0.50 (0.37, 0.70) | <0.001 |
|   Not informed | 4.477 | 901 | 733 | 2.843 | |
| Highest Temperature | 36.20 (36.00, 36.60) | 36.40 (36.00, 36.80) | 36.20 (36.00, 36.70) | 36.20 (36.00, 36.60) | <0.001 |
|   Not informed | 123 | 64 | 36 | 23 | |
| **Organ support at admission** | | | | | |
| Vasopressors or shock | 364 (3.3%) | 55 (3.3%) | 74 (3.1%) | 235 (3.4%) | 0.400 |
| Noninvasive ventilation | 311 (2.8%) | 79 (4.7%) | 69 (2.9%) | 163 (2.3%) | <0.001 |
| Mechanical ventilation | 86 (0.8%) | 28 (1.7%) | 16 (0.7%) | 42 (0.6%) | <0.001 |
| Renal replacement therapy | 60 (0.5%) | 14 (0.8%) | 23 (1.0%) | 23 (0.3%) | <0.001 |
| **Organ support during ICU stay** | | | | | |
| Noninvasive ventilation | 437 (4.0%) | 134 (8.0%) | 86 (3.6%) | 217 (3.1%) | <0.001 |
| Vasopressors or shock | 430 (3.9%) | 76 (4.5%) | 86 (3.6%) | 268 (3.8%) | 0.300 |
| Transfusion | 353 (3.2%) | 73 (4.3%) | 93 (3.9%) | 187 (2.7%) | <0.001 |
| Mechanical ventilation | 166 (1.5%) | 58 (3.4%) | 25 (1.1%) | 83 (1.2%) | <0.001 |
| Renal replacement therapy | 103 (0.9%) | 27 (1.6%) | 29 (1.2%) | 47 (0.7%) | <0.001 |
| High-flow nasal cannula | 23 (0.2%) | 6 (0.4%) | 4 (0.2%) | 13 (0.2%) | 0.400 |
| **Outcomes** | | | | | |
| Any complication during ICU Stay | 1,117 (10%) | 264 (16%) | 248 (10%) | 605 (8.7%) | <0.001 |
| ICU LOS, days | 3 (2, 4) | 3 (2, 4) | 3 (2, 4) | 3 (2, 5) | <0.001 |
| Hospital LOS, days | 5 (3, 7) | 5 (4, 7) | 4 (3, 6) | 5 (3, 7) | <0.001 |
| 60-day ICU Mortality | 109 (1.0%) | 33 (2.0%) | 18 (0.8%) | 58 (0.8%) | <0.001 |
| 60-day Hospital Mortality | 157 (1.4%) | 50 (3.0%) | 21 (0.9%) | 86 (1.2%) | <0.001 |

**Figure S1.** Flowchart of study inclusion

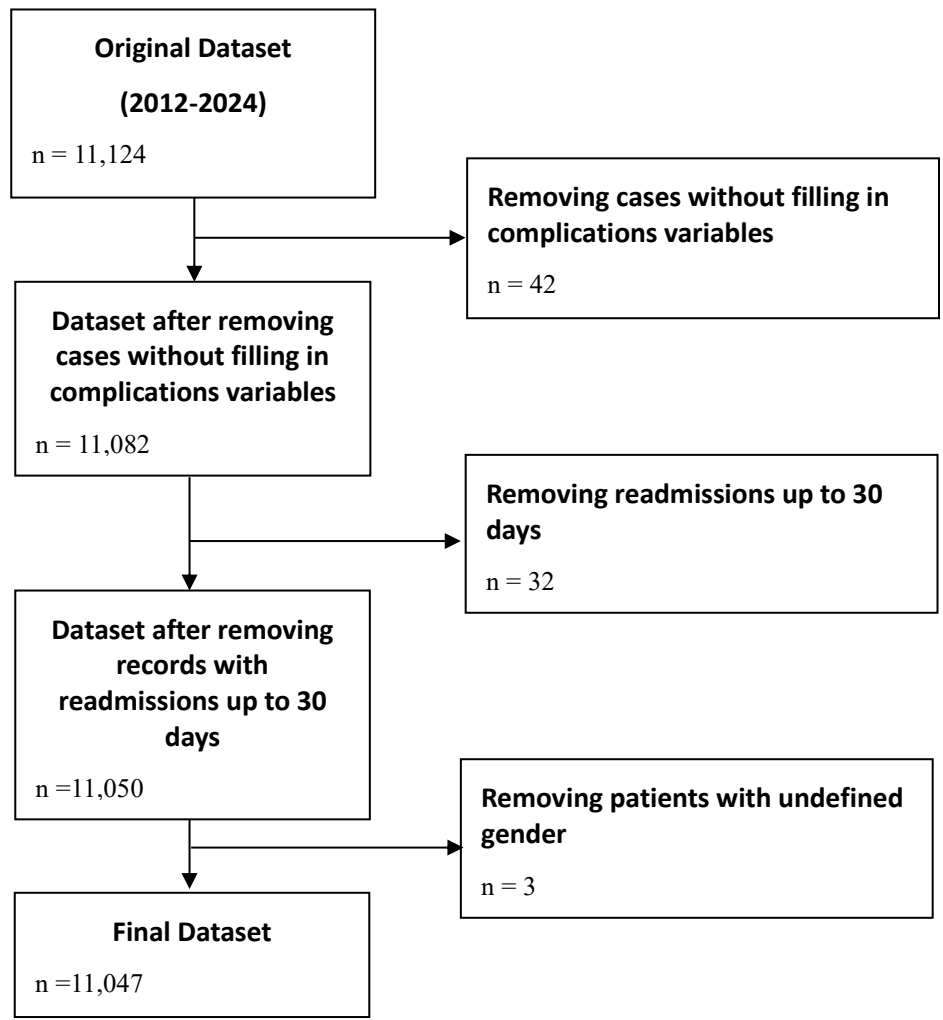

**Figure S2.** Distribution of platelets, leukocyte and lactate levels between two groups of patients (complicated and non-complicated patients). The blue curve represents patients who did not exhibit any complication while the red curve corresponds to complicated patients.

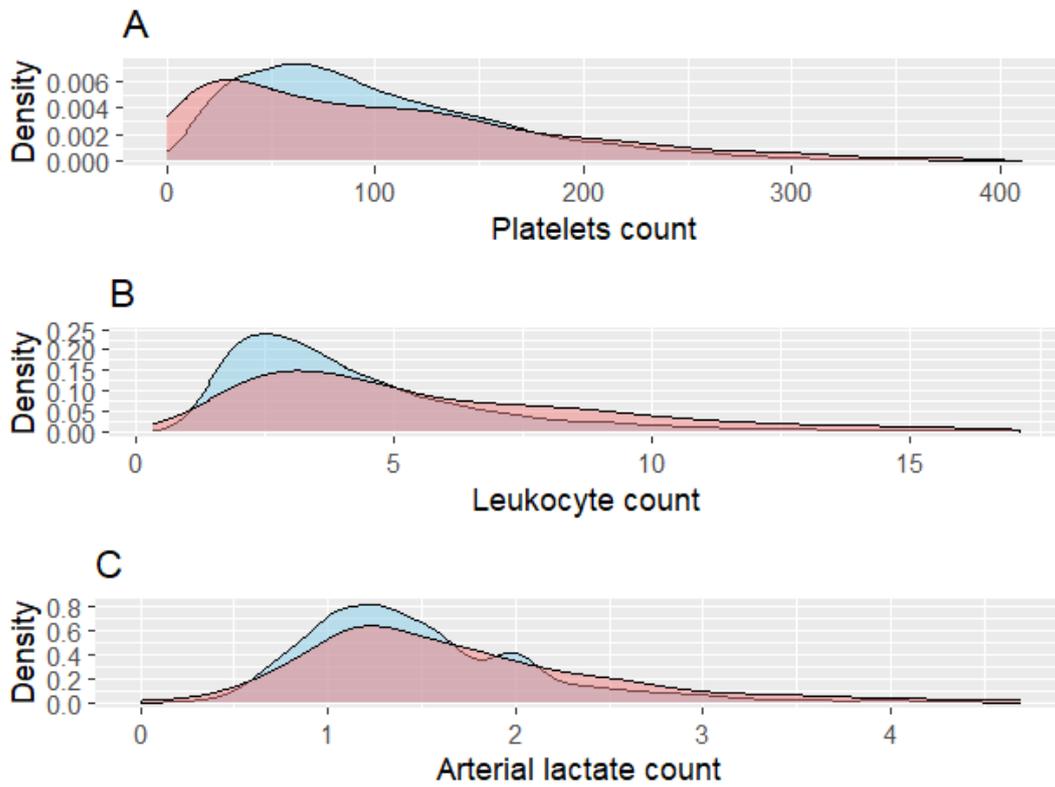

**Figure S3.** Correlation between model features

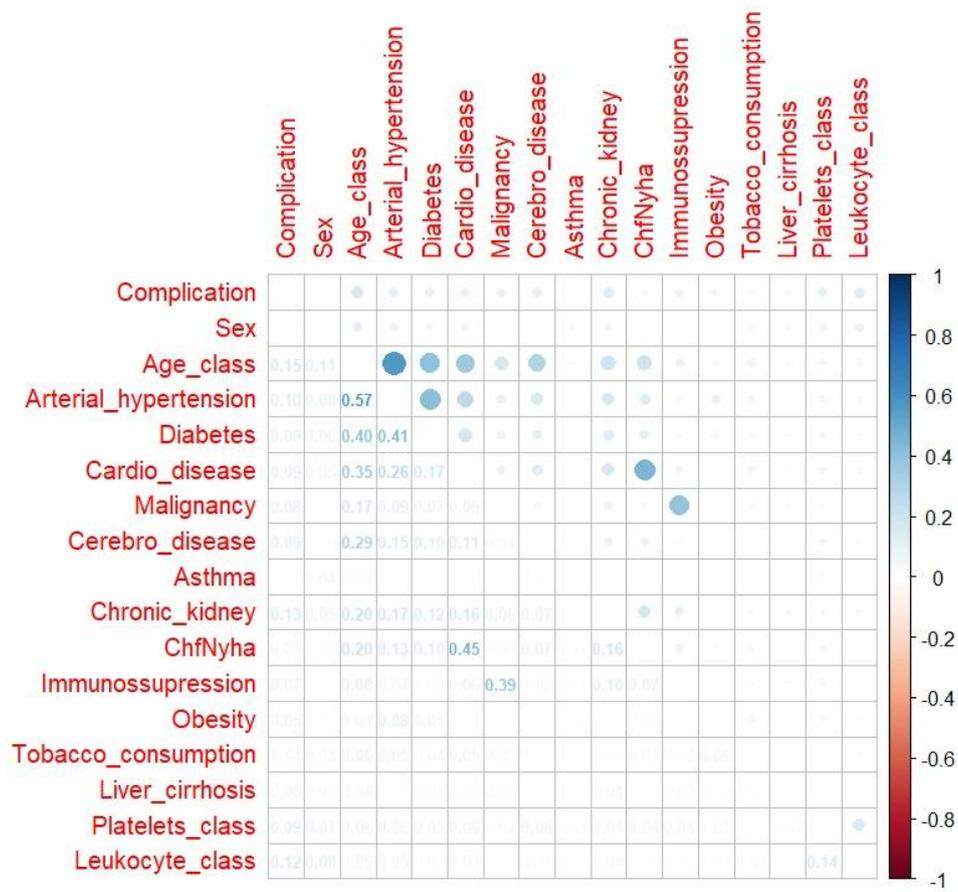

**Figure S4.** Calibration of each prediction model considering experiment (i)

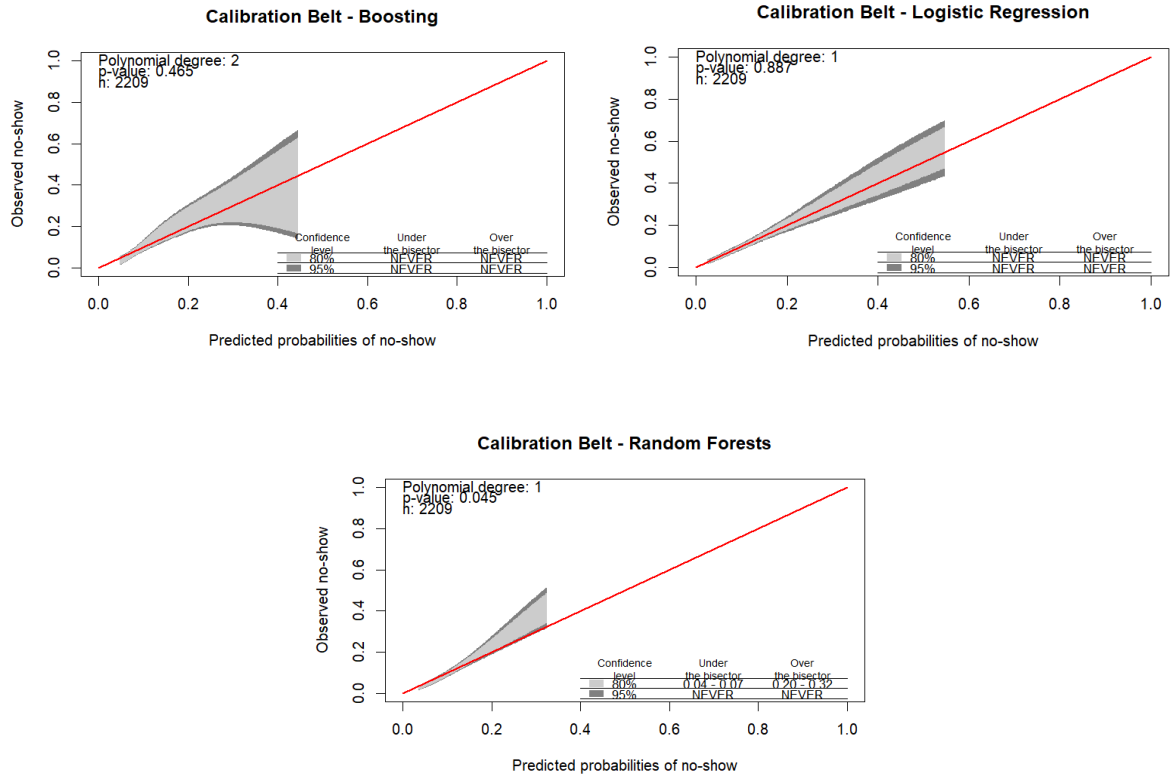

**Figure S5.** Calibration of each prediction model considering experiment (ii)

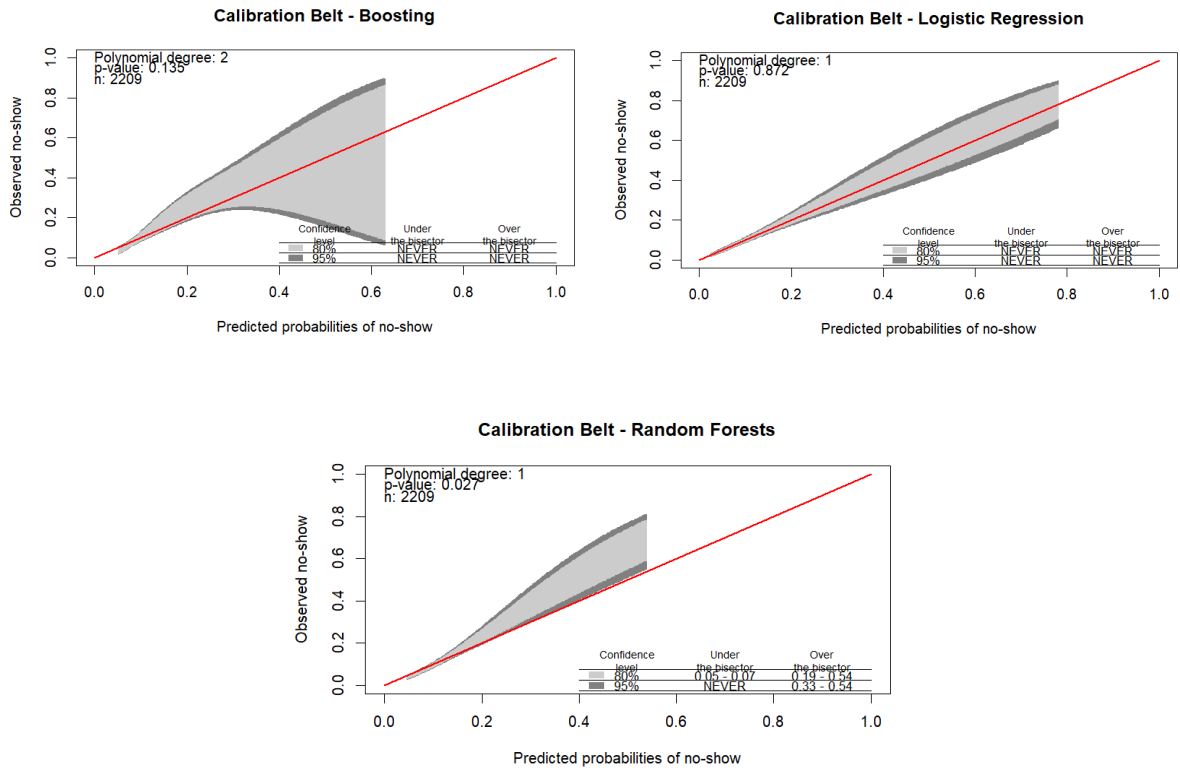